\documentclass[11pt,a4paper]{article}
\usepackage{booktabs}
\usepackage[hyperref]{emnlp2020}
\usepackage{times}
\usepackage{latexsym}

\usepackage{graphicx}
\usepackage{multirow}

\usepackage{microtype}

\usepackage{amsmath}
\aclfinalcopy 
\def\aclpaperid{102} 
\usepackage[utf8]{inputenc}
\usepackage{color,soul}
\definecolor{pink}{rgb}{1,0.8,0.8}
\definecolor{ltgreen}{rgb}{0.7,1,0.7}
\definecolor{ltblue}{rgb}{0.7,0.8,1}

\newcommand\blfootnote[1]{%
  \begingroup
  \renewcommand\thefootnote{}\footnote{#1}%
  \addtocounter{footnote}{-1}%
  \endgroup
}

\title{Mitigating Gender Bias in Machine Translation with Target Gender Annotations}

\author{Artūrs Stafanovičs*$^{\dagger\ddagger}$ \and Toms Bergmanis*$^{\dagger\ddagger}$ \and Mārcis Pinnis$^{\dagger\ddagger}$  \\ \\
  $^\dagger$Tilde / Vienības gatve 75A, Riga, Latvia \\
  $^\ddagger$Faculty of Computing, University of Latvia / Raiņa bulv.  19, Riga, Latvia\\ 
  {\tt \{firstname.lastname\}@tilde.lv}
}
\date{}

\begin{document}
\maketitle
\begin{abstract}
When translating ``\textit{The secretary asked for details.}'' to a language with grammatical gender, it might be necessary to determine the gender of the subject ``\textit{secretary}''. If the sentence does not contain the necessary information, it is not always possible to disambiguate. In such cases, machine translation systems select the most common translation option, which often corresponds to the stereotypical
translations, thus potentially exacerbating prejudice and marginalisation of certain groups and people. We argue that the information necessary for an adequate translation can not always be deduced from the sentence being translated or even might depend on external knowledge. 
Therefore, in this work, we propose to decouple the task of acquiring the necessary information from the task of learning to translate correctly when such information is available. 
To that end, we present a method for training machine translation systems to use word-level annotations containing information about subject's gender. 
To prepare training data, we annotate regular source language words with grammatical gender information of the corresponding target language words.
Using such data to train machine translation systems reduces their reliance on gender stereotypes when information about the subject's gender is available. Our experiments on five language pairs show that this allows improving accuracy on the WinoMT test set by up to 25.8 percentage points.
\vspace{-10pt}
\end{abstract}

\blfootnote{*First authors with equal contribution.}

\section{Introduction}
Most modern natural language processing (NLP) systems learn from natural language data. Findings of social sciences and corpus linguistics, however, indicate various forms of bias in the way humans use language \cite{coates1987women,butler1990feminism,fuertes2007corpus,rickford2016raciolinguistics}.
Thus the resulting NLP resources and systems also suffer from the same socially constructed biases, as well as inaccuracies and incompleteness \cite{jorgensen2015challenges,hovy2015tagging,prates2019assessing,vanmassenhove-etal-2019-lost,bordia-bowman-2019-identifying,davidson2019racial,NIPS2019_9479}.
Due to the prevalent use of NLP systems, their susceptibility to social biases becomes an increasingly significant concern as NLP systems not only reflect the biases learned but also amplify and perpetuate them further \cite{hovy-spruit-2016-social,crawford2017trouble,hleg2019ethics}.

This work concerns mitigating the manifestations of gender bias in the outputs of neural machine translation (NMT) systems in scenarios where the source language does not encode the information about gender that is required in the target language. An example is the translation of the English sentence ``\textit{The secretary asked for details.}'' into Latvian. In English, the gender of ``\textit{secretary}'' is ambiguous. In Latvian, however, there is a choice between the masculine noun ``\textit{sekretārs}'' and the feminine noun ``\textit{sekretāre}''. In cases when sentences do not contain the necessary information, NMT systems opt for translations which they have seen in training data most frequently.
Acquiring the necessary information, however, might require analysis of the text beyond the level of individual sentences or require incorporation of external knowledge. 

Falling back to biases, however, happens not only in the absence of the required information as NMT systems produce stereotyped translations even when clues about the subject's correct gender are present in the sentence \cite{stanovsky-etal-2019-evaluating}. This is in line with findings by \newcite{vanmassenhove-etal-2019-lost} who suggest that NMT systems produce biased outputs not only because of the biases present in data but also due to their tendency to exacerbate them.

To provide means for incorporation of external and explicit gender information, we propose a method for training NMT systems to use word-level gender annotations. To prepare training data, we project grammatical gender information of regular target language words onto the corresponding source language words. Albeit in some cases redundant, we expect that the grammatical gender information contains a useful learning signal that helps narrowing down the lexical choice of the correct target translation. As a result, the NMT system learns to rely on these annotations when and where they are available. In particular, in experiments on five language pairs, we show that the methods proposed here can be used in tandem with off-the-shelf co-reference resolution tools to improve accuracy on the WinoMT challenge set \cite{stanovsky-etal-2019-evaluating} by up to 25.8 percentage points.  \vspace{-4pt}

\subsection{Related work}
Recent recommendations for ethics guidelines for trustworthy AI recommend removing socially constructed biases at the source, the training data, prior to model training \cite{hleg2019ethics}. An example of work on debiasing training data is \newcite{zhao-etal-2018-gender} where authors identified sentences containing animate nouns and changed their grammatical gender to the opposite. \newcite{zmigrod-etal-2019-counterfactual} take it further by ensuring that not only the animate nouns but also the rest of the sentence is reinflected from masculine to feminine (or vice-versa), thus preserving the morpho-syntactic agreement of the whole sentence. 
The applicability of this line of work is still to be established as reinflecting sentences with co-references or pairs of parallel sentences in NMT pose an additional challenge.

A different take on addressing gender biases in NMT outputs is the work on alternative generation: given a gender-ambiguous source sentence and its translation, provide an alternative translation using the opposite gender. \newcite{habash-etal-2019-automatic} approach this as a gender classification and reinflection task for target language sentences to address the first person singular cases when translating from English into Arabic. 
\newcite{DBLP:journals/corr/abs-1811-01157} analyze trained NMT models to identify neurons that control various features, including gender information, that are used to generate the target sentence. 
In practice, however, such solutions are limited to simple source sentences where only one alternative in the target language is possible. 

A complementary approach is addressing gender bias in NMT as a problem of domain mismatch. When translating TED talks, \newcite{michel-neubig-2018-extreme} propose to adapt the NMT model for each speaker's attributes, thus also implicitly addressing previously poorly translated first-person singular cases.  \newcite{Saunders2020ReducingGB} describe methods for NMT model adaptation using a handcrafted gender-balanced dataset and a translation re-scoring scheme based on the adapted models. 

The closest line of work to ours is the work on the incorporation of external gender information in the NMT input. \newcite{Elaraby2018GenderAS} and \newcite{vanmassenhove-etal-2018-getting} prepend training data sentences with speaker gender information to improve spoken language translation when translating into languages with grammatical gender. \newcite{moryossef-etal-2019-filling} undertakes a similar approach at the inference time using phrases (e.g. \textit{``she said:''}) that imply the speaker's gender. The methods proposed in this work differ from the previous work in terms of annotation granularity: we propose to use token level annotations, while the previous work used one annotation per sentence. As our training data annotations are solely based on grammatical gender,  preparing them does not require any external gender information. Thus our approach is also simpler in terms of training data preparation compared to the previous work \cite{Elaraby2018GenderAS,vanmassenhove-etal-2018-getting}.\vspace{-5pt}

\paragraph{Social Impact}
We propose methods to mitigate the manifestations of gender bias in the outputs of NMT. Specifically, these methods provide explicit means to incorporate information about subjects referential or social gender in NMT, thus reducing gender-based stereotyping when translating into languages which encode for grammatical gender in animate nouns. An example of a use case and a beneficiary group is the translation of occupational nouns into languages which mark gender and people for whom stereotypes of their profession do not align with their gender. While these methods can relieve gender-based representational harms by reducing stereotyped translations, they, unfortunately,  provide no means for better representation of non-binary gender identities.  \vspace{-4pt}

\begin{figure*}
\centering
\includegraphics{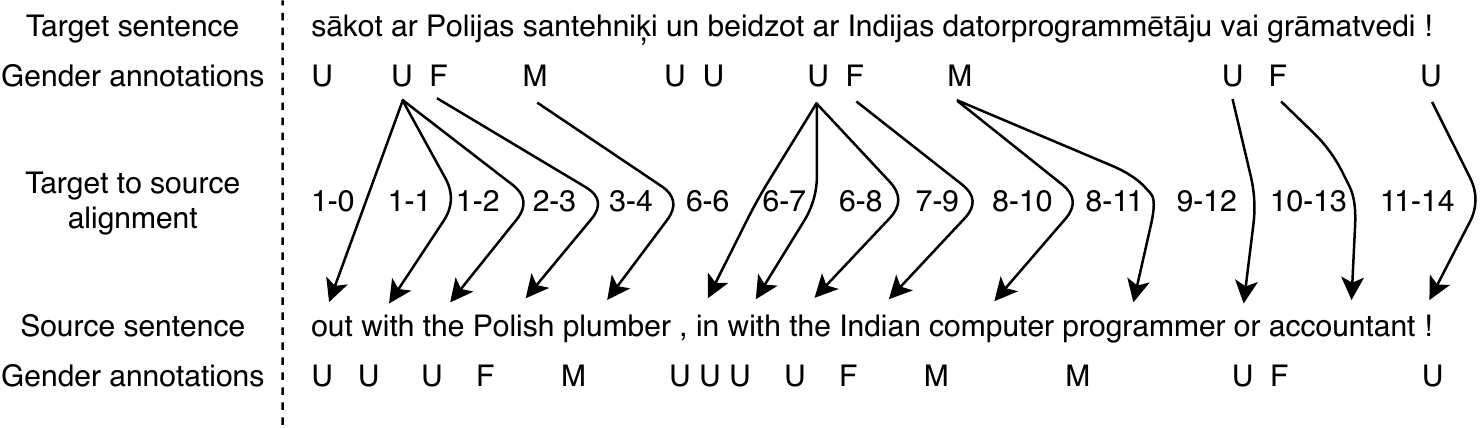}
\caption{Illustration of target to source projections of grammatical gender annotations. Sample sentences taken from the English-Latvian development set of the WMT2017 News Translation Task.}
\label{fig:gen-marking}
\end{figure*}

\section{Methods}\label{sec:methods}
When translating from languages without grammatical gender to languages with grammatical gender, certain words in the source sentence may not contain all the necessary information to produce an adequate and accurate translation. Examples are pronouns (e.g. \textit{I, me, they, them, themselves}), animate nouns such as job titles and proper nouns such as names and surnames, which depending on the sentence context can be ambiguous and consequently can be translated poorly. Previous work has also shown that NMT systems are better at translating sentences that align with socially constructed gender stereotypes because they are more frequently seen in training data \cite{stanovsky-etal-2019-evaluating,prates2019assessing}. 

To circumvent the degradation of NMT outputs due to 1) socially constructed biases and 2) absence of necessary information, we propose a method for training NMT systems to be aware of and use word-level target gender annotations (\textbf{TGA}). For training, we use data where regular source language words are annotated with the grammatical gender of their target language translations. We obtain such data by, first, morphologically tagging target language sentences to obtain information about their grammatical gender---\texttt{F} for feminine, \texttt{M} for masculine, \texttt{N} for neuter, and \texttt{U} for cases where grammatical gender is unavailable.  Then, we use word-level statistical alignments to project this information from the target language to the source language words (see Figure~\ref{fig:gen-marking} for an illustration). We use source-side factors \cite{sennrich-haddow-2016-linguistic} to integrate the projected annotations as an additional input stream of the NMT system. To ensure that the NMT systems are capable of producing adequate translations when gender annotations are not available---a frequently expected case at the test time---we apply TGA dropout. We do so by randomly replacing annotations for a random number of words with \texttt{U}.

While useful for animate nouns, such annotations might seem otherwise redundant because the majority of nouns in training data can be expected to be inanimate.  
However, for some inanimate nouns, the target language grammatical gender annotations can help narrowing down the lexical choice during training.
An example is the translation of \textit{``injury''} into Latvian, where ``\textit{injury}$\vert$\texttt{F}'' would result in ``\textit{trauma}'' while  ``\textit{injury}$\vert$\texttt{M}'' would correspond to ``\textit{ievainojums}''. Besides disambiguating animate nouns, annotations also disambiguate the grammatical gender of pronouns, proper nouns. Furthermore, grammatical gender annotations also concern adjectives and verbs, which in some languages have to agree in gender with the nouns they describe. Consequently, we expect that during training the NMT model will learn to use these annotations, as they contain valuable information about words in the target sentence. 

At inference time, we lean heavily on the observation that there the grammatical gender of animate nouns, pronouns, and proper nouns, and the intended referential gender coincide considerably. This is, however, a heuristic and not a rule (see \newcite{hellinger2015gender} for counterexamples).  Nevertheless, we assume that it is possible to use TGA in a referential sense of gender, thus injecting the NMT model with additional information about the subject's gender. Sources of such information can vary; in this paper, we show-case how to use TGA together with off-the-shelf co-reference resolution tools.

\begin{figure*}
\centering
    
\includegraphics{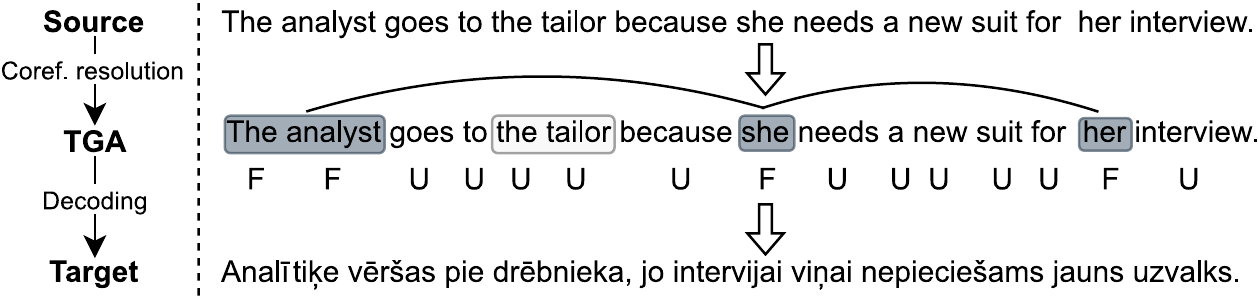}
\caption{WinoMT test suite translation process with TGA distilled from the output of automatic coreference resolution tool.}
\label{fig:tga-coreference}

\end{figure*}

\subsection{Evaluation: WinoMT Test Suite}
To measure the extent to which gender annotations reduce NMT systems' reliance on gender stereotypes, we use the WinoMT test suite \cite{stanovsky-etal-2019-evaluating}. WinoMT builds on the previous work on addressing gender bias in co-reference resolution by combining Winogender \cite{rudinger2018gender} and WinoBias \cite{zhao-etal-2018-gender} datasets in a test suite for automatic evaluation of gender bias in MT. All sentences in the WinoMT test set follow the Winograd Schema where anaphora resolution is required to find an antecedent for an ambiguous pronoun \cite{hirstanaphora}. In the case of datasets designed for evaluation of gender bias, the ambiguous pronoun refers to one of two entities which are referred to using titles of their professions. Professions and pronouns are chosen so that they either align with or diverge from the gender stereotypes  of each profession as reported by the U.S. Bureau of Labor Statistics \cite{zhao-etal-2018-gender}.

WinoMT tests if the grammatical gender of the translation of an antecedent matches the gender of the pronoun in the original sentence. Testing is done by morphologically analysing the target translation and aligning it with the source sentence. The WinoMT test suite scores MT outputs using multiple metrics:  \textbf{Accuracy} -- the percentage of correctly translated antecedents, \textbf{$\boldsymbol{\Delta G} $} -- difference in $F_1$ score between sentences with masculine and feminine antecedents, \textbf{$\boldsymbol{\Delta S}$} -- difference in accuracy between the set of sentences that either align with or diverge from the gender stereotypes of each profession. \newcite{Saunders2020ReducingGB} also propose to report \textbf{M:F} -- ratio of translations using masculine and feminine antecedents.

\section{Experimental Setting}
\paragraph{Languages and Data}
In all our experiments, we choose one source language without grammatical gender and five Indo-European languages in which nouns have grammatical gender (see Table~\ref{table:train-size}).
For all language pairs, we use training data from WMT news translation tasks. We do the necessary cleaning and filtering with Moses \cite{koehn-etal-2007-moses} pre-processing tools. 
To see how TGA is affected by data size, we also use much larger EN-LV proprietary data that we obtain from \href{https://www.tilde.com/products-and-services/data-library}{Tilde Data Libarary} by combining all EN-LV parallel corpora. The proprietary data are pre-processed using the Tilde MT platform \cite{pinnis-etal-2018-tilde}. 
Table~\ref{table:train-size} summarizes training data source and size statistics prior to adding TGA.
For all systems and language pairs, we use byte pair encoding (BPE) \cite{Gage:1994:NAD:177910.177914,sennrich-etal-2016-neural} to prepare joint source and target language BPE sub-word vocabularies. We use 30K BPE merge operations and use a vocabulary threshold of 50.

\paragraph{NMT Systems}
We use the default configuration of the Transformer \cite{NIPS2017_7181} NMT model implementation of the Sockeye NMT toolkit \citep{hieber-etal-2020-sockeye}. The exception is the use of source-side factors \cite{sennrich-haddow-2016-linguistic} with the dimensionality of 8 for systems using TGA, which changes the model's combined source embedding dimensionality from 512 to 520. We train all models using early stopping with patience of 10 based on their development set perplexity \cite{prechelt1998early}.

\begin{table}
\centering

\begin{tabular}{cccc}
\toprule
 & \textbf{Source} & \textbf{\# Sent.} & \textbf{News Test} \\
\midrule
EN-DE & WMT19 & 64.1M  & 2018\\
EN-FR & WMT15 & 39.1M  & 2015\\
EN-LV & Tilde & 22.7M & 2017 \\
EN-LV & WMT17 & 4.5M  & 2017 \\
EN-LT & WMT19 & 3.6M  & 2019\\
EN-RU & WMT17 & 25.0M  & 2015 \\
\bottomrule
\end{tabular}
\caption{Training data set source and size in millions of sentences prior to adding TGA.}\label{table:train-size} 

\end{table}

\begin{table}
\centering

\begin{tabular}{lcc}
\toprule
\textbf{Tagger} & \textbf{F1 masc.} & \textbf{F1 fem.}\\
\midrule
 \newcite{paikens2013morphological} & 98.6 & 98.7 \\
Stanza  & 94.7 & 95.1 \\
UDPipe  & 92.5 & 92.4 \\
\bottomrule
\end{tabular}

\caption{Performance of morphological taggers on gender feature classification evaluated on the Universal Dependencies test set.}\label{morph-f1}
\end{table}

\paragraph{Morphological Taggers} The preparation of training data with TGA and WinoMT evaluation relies on the outputs of a morphological tagger.  
If the tagger produces biased outputs, the TGA annotations might become too noisy to be useful. Furthermore, a biased morphological tagger could also render WinoMT evaluation unreliable.
Thus we first benchmark several morphological taggers on grammatical gender feature classification. We use Latvian as a development language because of the availability of lexicon-based and data-driven morphological analysis tools. Specifically, we use the Universal Dependencies
\footnote{\url{https://github.com/UniversalDependencies/UD_Latvian-LVTB}} test set to compare two data-driven tools -- the Stanza
toolkit \cite{qi-etal-2020-stanza} and UDPipe
\cite{udpipe:2017}. Additionally, we evaluate a dictionary-based morphological analyser and statistical tagger\footnote{\url{https://github.com/PeterisP/LVTagger}} by \newcite{paikens2013morphological}.
Table~\ref{morph-f1} gives F-1 scores on masculine and feminine feature tagging. Results indicate that none of the taggers exhibits salient bias in their tagging performance. As the only non-neural system yields better F-1 scores than the other two systems, we further compare Stanza and the tagger by \newcite{paikens2013morphological} in their impact on BLEU and WinoMT metrics. Results indicated that the choice of the tagger does not have a notable effect on BLEU scores. 
In terms of WinoMT accuracy scores, the NMT system that was trained using TGA prepared with Stanza yields an accuracy that is about 3\% better than the system using the tagger by \newcite{paikens2013morphological}. 
Thus, in all remaining experiments, we use the Stanza tagger as it provides pre-trained models for a wide range of languages.

\paragraph{TGA in Training Data}
Preparing training data with TGA requires statistical word alignments between words of source and target language sentences and a target language morphological tagger. To obtain word alignments, we use \textit{fast\_align}
\cite{dyer2013simple}. To obtain grammatical gender information of target language words, we use the Stanza morphological tagger. When training NMT systems with TGA, we combine two copies of the original training data: one where all source-side factors are set to \texttt{U} and the other containing TGA. 

\paragraph{TGA During Inference}

In training data, TGA annotate regular source language words with the grammatical gender information of corresponding target language words. We do not have access to the target language sentence during inference. Thus, we use co-reference resolution tools and extract the referential gender information from the source sentence instead. To do so, we first use co-reference resolution tools to obtain the co-reference graph. We then identify sub-graphs which contain gendered pronouns. Finally, we propagate the gender information within the graph and annotate the antecedents (see Figure~\ref{fig:tga-coreference}). We set the annotations for the remaining unannotated words to \texttt{U}.

We use neural co-reference resolution tools by AllenNLP
\footnote{\url{https://github.com/allenai/allennlp}} 
\cite{lee2017end} and  Hugging~Face\footnote{\url{https://github.com/huggingface/neuralcoref}} (based on work  by \newcite{clark2016deep}). We refer to these systems as \textbf{TGA AllenNLP} and \textbf{TGA HuggingFace} respectively.
We also report the performance of NMT with TGA, when TGA use oracle information directly taken from WinoMT datasets and refer to these as \textbf{TGA Oracle}.

\paragraph{Evaluation}
We evaluate general translation quality using the BLEU \cite{papineni-etal-2002-bleu} metric evaluated over WMT test sets. To calculate BLEU, we use SacreBLEU\footnote{SacreBLEU hash: \url{BLEU+case.mixed+numrefs.1+smooth.exp+tok.13a+version.1.3.6}} \cite{post-2018-call} on cased, detokenized data. Reference test sets are only pre-processed using Moses  punctuation normalization script\footnote{\url{https://github.com/moses-smt/mosesdecoder/blob/master/scripts/tokenizer/normalize-punctuation.perl}}. 
We use the WinoMT test suite \cite{stanovsky-etal-2019-evaluating} to measure gender bias of our NMT systems.

\section{Results and Discussion}
\begin{table}
\centering
\small{
\begin{tabular}{clcccc}
\toprule

 & \multicolumn{5}{c} {\textbf{WMT Data Systems }} \\

                        &              & \textbf{Acc.}    & \textbf{$\Delta$ G }     & \textbf{$\Delta$ S}    & \textbf{M:F }\\ 
\midrule
\multirow{4}{*}{\rotatebox[origin=c]{90}{EN-DE}} & \textbf{Baseline}     & 66.7 & 10.2 & 14.4   & 2.6 \\  
& \textbf{TGA Oracle }& 89.0 & -4.7 & 1.7    & 1   \\
                       & \textbf{TGA HuggingFace} & 77.6  & -0.1  & 11.9 & 1.6 \\ 
                       & \textbf{TGA AllenNLP} &   81.5      &  -2.0       &  11.1      &   1.4  \\ 
                        \midrule
\multirow{4}{*}{\rotatebox[origin=c]{90}{EN-FR}} & \textbf{Baseline}     & 48.6  & 29.8  & 11.8 & 5.5 \\ 
& \textbf{TGA Oracle} & 81.5  & 1.4   & 2.8  & 1.2 \\
                       & \textbf{TGA HuggingFace}& 67.8  & 4.9   & 12.4 & 2   \\ 
                       & \textbf{TGA AllenNLP} &   74.4      &  1.6       &  10.1      &   1.6  \\ 
                        \midrule
\multirow{4}{*}{\rotatebox[origin=c]{90}{EN-LV}} & \textbf{Baseline}     & 27.9  & 26.0  & 9.6  & 3.9 \\ 
                       & \textbf{TGA Oracle} & 42.7  & 15.9  & 10.3 & 2.9 \\ 
                       & \textbf{TGA HuggingFace} & 38.6  & 19.7  & 18.1 & 3.0 \\ 
                       & \textbf{TGA AllenNLP} &   39.3      &  18.1       &  18.6      &   2.8  \\ 
 \midrule 
\multirow{4}{*}{\rotatebox[origin=c]{90}{EN-LT}} & \textbf{Baseline}     & 38.0  & 32.6  & 6.5  & 5.9 \\
                       & \textbf{TGA Oracle} & 52.8  & 15.2  & 4.0  & 2.7   \\
                       & \textbf{TGA HuggingFace} &   43.4      &  22.5       &  7.6      &   3.9  \\
                       & \textbf{TGA AllenNLP} &   47.2      &  17.7       &  5.1      &   3.1  \\ 
 \midrule
\multirow{4}{*}{\rotatebox[origin=c]{90}{EN-RU}} & \textbf{Baseline}     & 32.3  & 37.7  & 14.1 & 8.4 \\  
                       &\textbf{TGA Oracle} & 55.9  & 10.6   & 14.0 & 2.5 \\
                       & \textbf{TGA HuggingFace} & 45.4  & 24.8  & 13.7 & 4.4 \\ 
                       & \textbf{TGA AllenNLP} &   51.4      &  17.0       &  15.2      &   3.2  \\ 
 \midrule
 & \multicolumn{5}{c} {\textbf{Proprietary Large Data System }} \\ 
                        &              & \textbf{Acc.}    & \textbf{$\Delta$ G }     & \textbf{$\Delta$ S}    & \textbf{M:F }\\ 
\midrule
\multirow{4}{*}{\rotatebox[origin=c]{90}{EN-LV}} & \textbf{Baseline}      & 42.0 & 27.9 & 16.6 & 4.9 \\                        & \textbf{TGA Oracle} & 55.1  & 4.8   & 18.2 & 1.7 \\ 
                       & \textbf{TGA HuggingFace} & 46.2  & 13.5  & 24.1 & 2.6 \\ 
                       & \textbf{TGA AllenNLP} &   49.9      &  10.8       &  23.1      &   2.3  \\ 

\bottomrule
\end{tabular}
}
\caption{Results on WinoMT test suite.} \label{table:WinoMT}
\end{table}

Results from experiments evaluating gender bias using the WinoMT test suite are provided in Table~\ref{table:WinoMT}. First, we observe that all baseline systems show a strong bias towards generating translations using masculine forms. The EN-RU baseline system is the most biased as it produces only one translation hypothesis with a feminine antecedent for every 8.4 hypotheses containing masculine antecedents. Meanwhile the EN-DE baseline system is the least biased with the M:F ratio being much lower -- 2.6 (see the last column of Table~\ref{table:WinoMT}). Our baseline systems for EN-DE, EN-FR and EN-RU language pairs, however, show comparable $\Delta G$ and WinoMT accuracy results to those reported by \newcite{stanovsky-etal-2019-evaluating} for several publicly available commercial systems. These results confirm that our baselines, although being strongly biased, are not unordinary.

Results from experiments using TGA with oracle gender information show an improvement in WinoMT accuracy and $\Delta G$ for all language pairs (see Table~\ref{table:WinoMT} TGA Oracle). These results demonstrate that when training MT systems to use TGA reduces their reliance on gender stereotypes when information about the subject's gender is available, proving the usefulness of methods proposed here. Despite the availability of oracle gender information, none of the systems is entirely bias-free or obtains 100\% accuracy. Thus methods proposed here could be combined with others, such as those proposed by \newcite{Saunders2020ReducingGB}, to achieve further improvements.

\paragraph{Effect on BLEU} As expected, using TGA with reference sentence grammatical gender annotations has a positive effect on BLEU, thus confirming our hypothesis why and how the NMT system learns to rely on TGA as an additional source of information during training (see Table~\ref{table:bleu}).  It is equally important, however,  that, when training NMT systems to use TGA, it does not degrade their performance when gender information is not necessary or is unavailable. Thus we test our systems for such cases by setting all TGA values to \texttt{U} and compare them to the baseline systems (see Table~\ref{table:bleu}). To test for statistically significant differences between the results of NMT systems we use pairwise bootstrap re-sampling \cite{koehn2004statistical} and significance threshold of $0.05$. 
Results indicate no statistically significant differences between systems using uninformative TGA values and their baseline counterparts with an exception of results for EN-RU systems ($\Delta0.4$ BLEU), which we find to be statistically significant.

\paragraph{Effect of Data Size}
To analyze gender bias and TGA performance depending on the quality and size of the training data, we use much larger EN-LV proprietary data  (see Table~\ref{table:train-size})     to train production-grade NMT systems and contrast them with EN-LV WMT data systems (see the two EN-LV sections in Table~\ref{table:WinoMT} and Table~\ref{tab:lv-systems}). 
First of all, we notice that although the large data baseline has higher WinoMT accuracy than the WMT data system, it has a similar $\Delta G$. 
Decomposing $\Delta G$ as male and female grammatical gender F-1 scores (Table~\ref{tab:lv-systems}), however,  clarifies that, although similarly skewed, the large data baseline has higher F-1 scores than the WMT data baseline. 
Next, we note, that larger training data size has a positive effect on the system's ability to use TGA more effectively as the large data system using TGA has a greater improvement on the two metrics measuring bias -- $\Delta G$ and M:F\footnote{$\Delta S$ results are not reliable or comparable when M:F ratios are large or differ by a large value. See result section of \newcite{Saunders2020ReducingGB} for more discussion.} than its WMT data counterpart relative to its baseline. 
These findings suggest that TGA is a method that is applicable not only in small data settings but also in large data settings, such as commercial systems, for which it is even more effective.

\paragraph{Plugging-in Co-reference Resolution Tools}  Finally, we experiment with TGA using gender information provided by two off-the-shelf co-reference resolution tools, AllenNLP and Hugging Face. Results show that using TGA with either of the tools outperforms baseline systems for all languages pairs. Furthermore, TGA with gender information provided by AllenNLP shows only a 4.5 to 7.1\% drop in WinoMT accuracy compared to results when using TGA with oracle information.  To put this in perspective, \newcite{Saunders2020ReducingGB} required a handcrafted gender-balanced profession set and additional rescoring models, for their EN-DE system to obtain comparable WinoMT accuracy and  $\Delta G$ without loss of translation quality. In contrast, the methods proposed here require tools that are readily available, making them easily applicable in practice.

\begin{table}
\centering
\begin{tabular}{lc|cc}
\\ \toprule

               &\textbf{Basline}                          & \textbf{ TGA}     & \textbf{All TGA= \texttt{U}}     \\ \midrule

\textbf{EN-DE} & 45.4                               & 49.5   & 45.3               \\
\textbf{EN-FR} & 36.6                               & 40.9     & 36.4             \\
\textbf{EN-LV} & 16.6                                  & 18.9   & 17.0              \\
\textbf{EN-LT} & 14.8                                & 16.6    & 14.7             \\
\textbf{EN-RU} & 27.1                              & 31.6    & 26.7               \\ \bottomrule
\end{tabular}
\caption{Comparison of test set performance measured in BLEU for Baseline systems and systems trained using TGA. TGA: performance when using reference sentence \emph{grammatical gender} annotations. All TGA=\texttt{U}: performance when all annotations set to be unknown.}\label{table:bleu}
\end{table}

\begin{table}
\centering
\small
\setlength{\tabcolsep}{5pt}
\begin{tabular}{lccc|ccc}
\toprule
                     & \multicolumn{3}{c|}{\textbf{Male}} & \multicolumn{3}{c}{\textbf{Female}} \\
 & \textbf{F-1} & \textbf{P} & \textbf{R} & \textbf{F-1} & \textbf{P} & \textbf{R} \\
 \midrule
\multicolumn{1}{l}{} & \multicolumn{6}{c}{\textbf{WMT Data System}}                                     \\
\midrule
\textbf{Baseline}    & 47.2      & 48.3      & 46.2      & 21.2       & 53.9       & 13.2      \\
\textbf{TGA Oracle}  & 58.5      & 56.0      & 61.3      & 42.5       & 74.7       & 29.7      \\
\midrule
\multicolumn{1}{l}{} & \multicolumn{6}{c}{\textbf{Proprietary Large Data System}}                       \\
\midrule
\textbf{Baseline}    & 58.8      & 50.8      & 69.7      & 30.9       & 70.3       & 19.8      \\
\textbf{TGA Oracle } & 66.9      & 65.8      & 68.0      & 62.1       & 83.3       & 49.5     \\
\bottomrule
\end{tabular}
\caption{Results of antecedent translation. Reporting grammatical gender F-1 score, precision (P) and recall (R) for EN-LV systems trained on WMT and proprietary large data. }
\label{tab:lv-systems}
\end{table}

\section{Conclusions}

We proposed a method for training MT systems to use word-level annotations containing information about the subject's gender. To prepare training data, the method requires a morphological tagger to annotate regular source language words with grammatical gender information of the corresponding target language words. During inference, annotations can be used to provide information about subjects' referential or social gender obtained by analyzing text beyond sentence boundaries or externally. In experiments with five language pairs, we showed that using such gender annotations reduces NMT systems' reliance on gender stereotypes in principle. We then further showed one way for how these findings can be used in practice by using off-the-shelf co-reference resolution tools. 

The method proposed here decouples the task of acquiring the necessary gender information from the task of learning to translate correctly when such information is available. Thus system's ability to use such information can be achieved independently from its availability at training time. This allows for application-specific sources of gender information. Examples are the translation of chat or social media content, where users may choose to indicate their gender or translation of whole documents, where gender information may be obtained using annotations and anaphora resolution. Thus, we believe that the methods proposed here, will provide means to limit the propagation of gender stereotypes by NMT systems when translating into languages with grammatical gender. 

The source code to reproduce our results for the publicly available data sets is published on GitHub\footnote{\url{https://github.com/artursstaf/mitigating-gender-bias-wmt-2020}}.

\section*{Acknowledgements}
This research was partly done within the scope of the undergraduate thesis project of the first author at the University of Latvia and supervised at Tilde. 

This research has been supported by the European Regional Development Fund within
the joint project of SIA TILDE and University of Latvia “Multilingual Artificial Intelligence Based Human Computer Interaction” No. 1.1.1.1/18/A/148.

\bibliography{references_bakalaurs,references_literatura_review,toms}
\bibliographystyle{acl_natbib}

\end{document}